\documentclass{article}
\usepackage{spconf,amsmath,graphicx,hyperref}
\usepackage{amssymb}  
\usepackage{multirow}  
\usepackage{booktabs} 
\usepackage{graphicx}  
\usepackage{makecell}

\title{Taming the Light: Illumination-Invariant Semantic 3DGS-SLAM}
\name{Shouhe Zhang\textsuperscript{1,2}, Dayong Ren\textsuperscript{3,*}, Sensen Song\textsuperscript{1,2,*}, Yurong Qian\textsuperscript{1,2} and Zhenhong Jia\textsuperscript{1,4} \thanks{This work was supported in part by the National Natural Science Foundation of China under Grant No. 62261053, the Project of Science and Technology Department of Xinjiang Uygur Autonomous Region under Grant 2024D01C240, the Tianshan Innovation Team Program of Xinjiang Uygur Autonomous Region of China (2023D14012).}}
\address{\textsuperscript{1}School of Computer Science and Technology, Xinjiang University\\
\textsuperscript{2}Joint International Research Laboratory of Silk Road Multilingual Cognitive Computing\\
Xinjiang University, Urumqi Xinjiang, 830046, China\\
\textsuperscript{3}National Key Laboratory for Novel Software Technology, Nanjing University, Nanjing 210023, China.\\
\textsuperscript{4}Key Laboratory of Signal Detection and Processing, Xinjiang University, Urumqi, 830046, China
}
\begin{document}
\ninept

\maketitle
\begin{abstract}
Extreme exposure degrades both the 3D map reconstruction and semantic segmentation accuracy, which is particularly detrimental to tightly-coupled systems. To achieve illumination invariance, we propose a novel semantic SLAM framework with two designs. First, the Intrinsic Appearance Normalization (IAN) module proactively disentangles the scene's intrinsic properties, such as albedo, from transient lighting. By learning a standardized, illumination-invariant appearance model, it assigns a stable and consistent color representation to each Gaussian primitive. Second, the Dynamic Radiance Balancing Loss (DRB-Loss) reactively handles frames with extreme exposure. It activates only when an image's exposure is poor, operating directly on the radiance field to guide targeted optimization. This prevents error accumulation from extreme lighting without compromising performance under normal conditions. The synergy between IAN's proactive invariance and DRB-Loss's reactive correction endows our system with unprecedented robustness. Evaluations on public datasets demonstrate state-of-the-art performance in camera tracking, map quality, and semantic and geometric accuracy.
\end{abstract}

\section{Introduction}

Recent advancements in neural radiance fields have revolutionized 3D scene representation, with 3D Gaussian Splatting (3DGS) emerging as a particularly potent technique\cite{kerbl20233d}. Its ability to achieve high-fidelity, real-time rendering from sparse image sets has unlocked new possibilities in robotics and augmented reality, most notably in the domain of Simultaneous Localization and Mapping (SLAM). By representing scenes as a collection of explicit 3D Gaussians, 3DGS-based SLAM systems\cite{splatam,monogs} can construct photorealistic maps while simultaneously tracking camera pose\cite{guo2024lidar,li2024dl,li2023online}, paving the way for creating true-to-life digital twins of our environment\cite{kerbl20233d,panoptic-neural-fields}.

However, the transition from controlled datasets to the unpredictable conditions of the real world exposes a critical vulnerability in these systems: the lack of robustness to illumination variations. The foundational assumption of photometric consistency—that a scene point maintains a constant appearance across different views—is frequently violated in practice\cite{NeRF-and-Gaussian-Splatting-SLAM,Online-photometric-calibration,real-time}. Drastic changes in lighting, such as moving from shadow to direct sunlight, or the automatic exposure adjustments of a moving camera, can cause identical scene elements to appear dramatically different.This variance corrupts the optimization process, leading to geometric artifacts, color inconsistencies in the map, and even catastrophic tracking failure\cite{Nerf-slam,Mip-nerf-360,semanticfusion}. This challenge is further magnified in the context of semantic SLAM\cite{Sni-slam,Sgs-slam,Hier-slam,song2024adaptive,zhang2025fuzzy,ren2017practical}, where the goal is not only to map the world but also to understand it. In coupled systems where geometry and semantics are mutually beneficial, illumination variance wreaks havoc on both fronts. The same surface under different exposures can yield vastly different feature representations, confusing semantic segmentation\cite{chen2024geosegnet,ren2022point,ren2024spiking,diao2025zigzagpointmamba,ren2017practical,ren2022sae} networks and leading to erroneous labeling\cite{semanticfusion,maskfusion,kimera}. This, in turn, feeds incorrect semantic priors back into the geometric reconstruction, creating a vicious cycle of degradation that undermines the entire system\cite{semanticfusion,nerflets}.

To address these fundamental limitations, we propose a novel illumination-invariant semantic SLAM framework designed to "tame" the effects of real-world lighting. Our approach is built on a two-pronged strategy: a proactive module that standardizes scene appearance at its core, and a reactive mechanism that dynamically compensates for extreme exposure variations during optimization. Our main contributions are centered around two key designs:

First, we introduce the Intrinsic Appearance Normalization (IAN) module, which proactively disentangles the scene's intrinsic properties from transient lighting. Instead of learning continuous colors that are easily influenced by lighting, our module constrains the appearance of each Gaussian to a discretized, canonical color palette. This quantization acts as a powerful regularizer, forcing the model to learn a stable, underlying albedo for scene surfaces, thereby achieving a standardized and illumination-invariant appearance representation across the entire map. Second, to reactively handle severe, per-frame brightness shifts, we propose the Dynamic Radiance Balancing Loss (DRB-Loss). This component learns to model per-image exposure variations through a set of latent parameters. Crucially, our loss function is adaptive and structure-aware; it applies this photometric correction only when it detects a significant structural inconsistency between the rendered and the real image. This dynamic behavior ensures that the system robustly compensates for challenging exposure without using the correction mechanism to mask underlying geometric errors, leading to more stable and accurate optimization. By integrating these proactive and reactive mechanisms, our framework achieves an unprecedented level of robustness. Comprehensive evaluations on challenging public datasets demonstrate that our method delivers state-of-the-art performance in camera tracking, map reconstruction quality, and semantic and geometric accuracy, successfully navigating the complex and varied illumination conditions of real-world environments.

\section{METHOD}
\label{sec:method}

\begin{figure}[t!]
\centering

\includegraphics[scale=0.25, trim=0cm 3cm 0cm 1.5cm, clip]{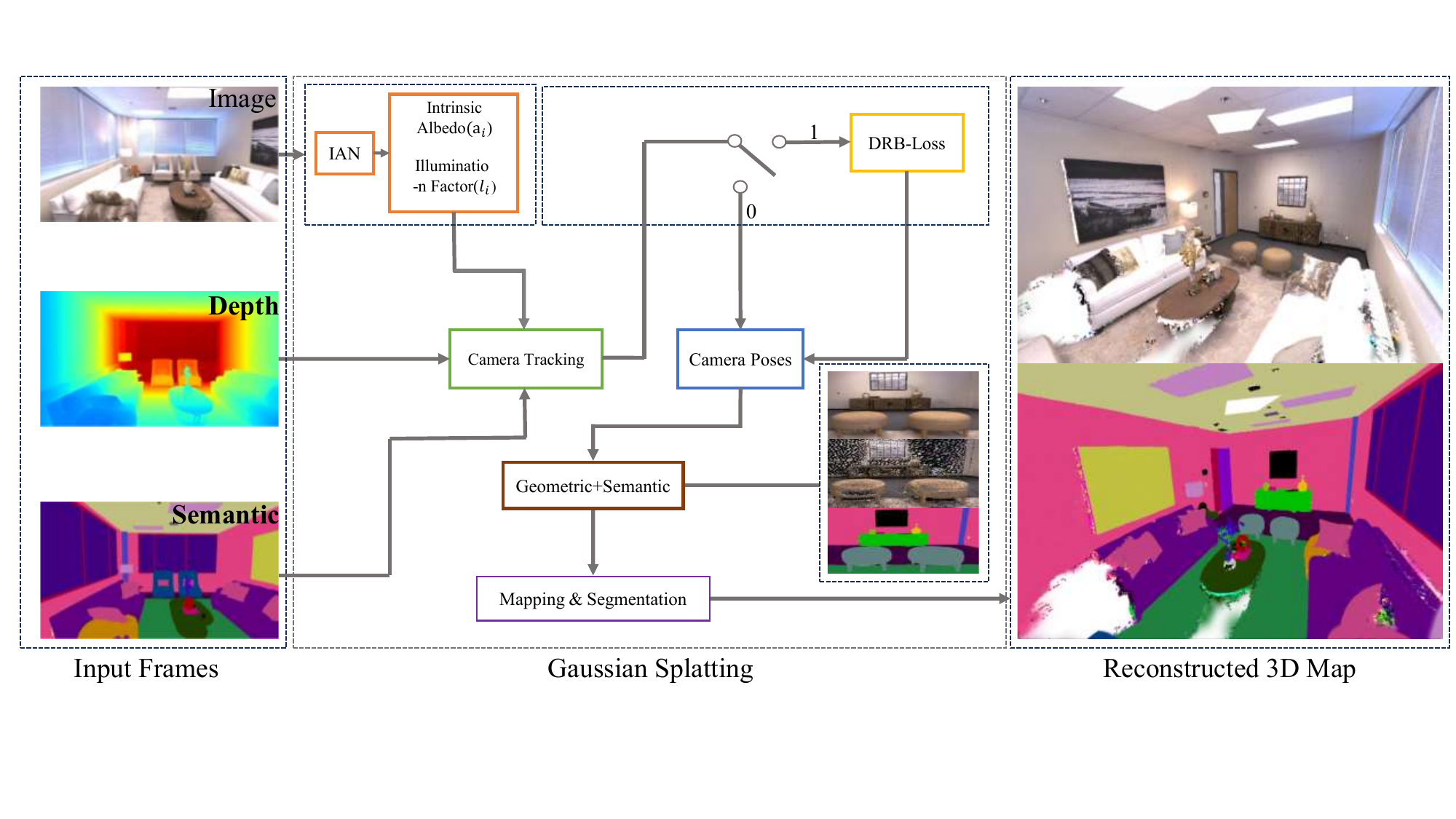}
\caption{The framework processes input frames by first disentangling albedo and illumination with the IAN module. It then performs camera tracking, using the DRB-Loss to robustly handle exposure variations. Finally, the system jointly optimizes a dense, high-fidelity map containing both geometric and semantic information.}
\label{fig:framework}
\end{figure}

In this section, we detail our illumination-invariant semantic 3DGS-SLAM framework. The core principle of our method is to \textbf{disentangle} the scene's appearance into two components: a stable \textbf{intrinsic albedo} and a varying \textbf{transient illumination}. Based on this decomposition, we introduce two key modules: the \textbf{Intrinsic Appearance Normalization (IAN)} module, which proactively learns a canonical color representation for the scene, and the \textbf{Dynamic Radiance Balancing Loss (DRB-Loss)}, which reactively compensates for extreme illumination shifts. By integrating these modules, our system jointly optimizes geometry, appearance, and semantics during tracking and mapping, ultimately reconstructing a high-fidelity, illumination-robust 3D semantic map. The overall pipeline is illustrated in Fig.~\ref{fig:framework}.

\subsection{Differentiable Gaussian Representation and Rendering}

To enable end-to-end optimization, we first define our scene representation based on 3D Gaussians. Each Gaussian primitive $G_i$ is parameterized by a set of attributes:
\begin{equation}G_i=\{\mu_i,r_i,\sigma_i,a_i,l_i,s_i\}.\end{equation}
where $\boldsymbol{\mu}_i \in \mathbb{R}^3$ is the center, $\boldsymbol{r}_i \in \mathbb{SO}(3)$ is the rotation, and $\sigma_i \in \mathbb{R}$ is the opacity.

Critically, deviating from standard 3DGS, we explicitly model the final appearance color $\boldsymbol{c}_i$ as the element-wise product of an intrinsic albedo $\boldsymbol{a}_i \in [0, 1]^3$ and a transient illumination factor $\boldsymbol{l}_i \in \mathbb{R}^+$. This relationship, $\boldsymbol{c}_i = \boldsymbol{a}_i \odot \boldsymbol{l}_i$, is central to our framework. Additionally, $\boldsymbol{s}_i$ represents the semantic label of the Gaussian. These parameters are optimized via a differentiable rendering pipeline.

\subsubsection{Color, Depth, and Semantic Rendering}
Following the standard 3DGS pipeline \cite{kerbl20233d}, we project the 3D Gaussians onto the 2D image plane and synthesize the final pixel attributes using volumetric $\alpha$-blending. For a given pixel $p$, the rendered attributes are computed as follows:

\textbf{Color Rendering.} The final color $C(p)$ is a blend of each Gaussian's appearance $\boldsymbol{c}_i$, which is the product of its intrinsic albedo and illumination factor.
\begin{equation}
C(p)=\sum_{i=1}^N \boldsymbol{c}_i \cdot f_i(p) \cdot \prod_{j<i}(1-f_j(p)).
\end{equation}
where $f_i(p)$ is the influence weight of Gaussian $i$ on pixel $p$, determined by its projected 2D covariance and opacity $\sigma_i$.

\textbf{Depth Rendering.} The pixel depth $D(p)$ is similarly computed by blending the depth $d_i$ of each Gaussian center in the camera coordinate system, providing geometric supervision.
\begin{equation}
D(p)=\sum_{i=1}^N d_i \cdot f_i(p) \cdot \prod_{j<i}(1-f_j(p)).
\end{equation}

\textbf{Semantic Rendering.} To facilitate joint semantic-geometric optimization, the semantic map $S(p)$ is rendered by blending the semantic labels $\boldsymbol{s}_i$ of the Gaussians in a manner consistent with color rendering \cite{Sgs-slam}.
\begin{equation}
S(p)=\sum_{i=1}^N \boldsymbol{s}_i \cdot f_i(p) \cdot \prod_{j<i}(1-f_j(p))
.\end{equation}

\subsection{Intrinsic Appearance Normalization (IAN)}

Directly optimizing continuous RGB colors allows the model to bake transient lighting effects into the scene's color representation, leading to inconsistencies and artifacts in the reconstructed map. To address this, we introduce the IAN module, which employs \textbf{color quantization as a strong regularizer} to compel the network to learn a stable and canonical intrinsic albedo $\boldsymbol{a}_i$ for scene surfaces, independent of lighting conditions. We discretize the continuous albedo space into a fixed, canonical palette. Specifically, during optimization, we enforce that each RGB channel of the intrinsic albedo $\boldsymbol{a}_i$ (normalized to $[0, 1]$) maps to one of four discrete values. This is achieved by applying the following rule:
\begin{equation}a_v^{\prime}=
\begin{bmatrix}
4\cdot a_v
\end{bmatrix}\cdot0.25+0.125 .\end{equation}
where $a_v$ is the original continuous channel value and $a_v'$ is its quantized counterpart. This rule constrains each albedo channel to the set $\{0.125, 0.375, 0.625, 0.875\}$. By forcing the albedo into this standardized representation, the model must attribute illumination variations to the dedicated illumination factor $\boldsymbol{l}_i$, thereby achieving a normalized and robust appearance representation across the map.

\begin{figure}[h!]
\centering
\includegraphics[scale=0.25,  trim=0cm 0cm 0cm 0cm, clip]{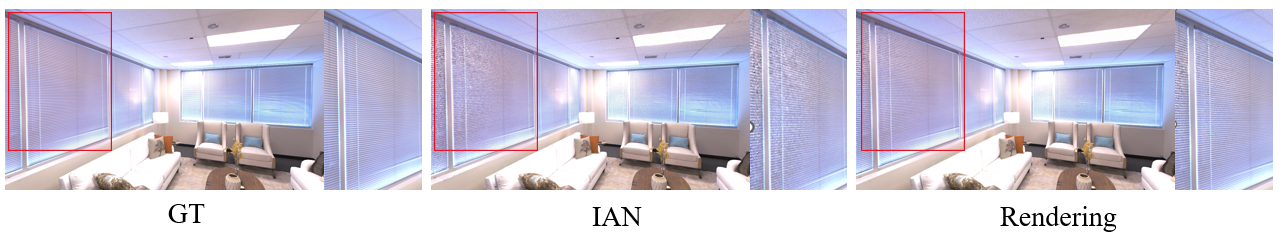}
\caption{IAN module's effect on intrinsic appearance.}
\label{fig:INA}
\end{figure}
\vspace{-15pt}
\subsection{Dynamic Radiance Balancing Loss (DRB-Loss)}

While the IAN module provides a stable base appearance for the map, it may not be sufficient to handle drastic, frame-wide radiance changes caused by factors like camera auto-exposure or moving between vastly different lighting environments. We, therefore, design the DRB-Loss as a reactive mechanism that activates only when necessary to model and compensate for these severe per-frame photometric variations. We dynamically detect the need for exposure correction using the Structural Similarity Index (SSIM) between the rendered image $I_{\mathrm{render}}$ and the ground-truth image $I_{\mathrm{gt}}$. When drastic illumination shifts occur (e.g., over or underexposure), the image's local structure (edges, textures) deviates significantly from that of the map rendered under a stable lighting assumption, leading to a drop in the SSIM score.
\begin{equation}
S = \mathrm{SSIM}(I_{\mathrm{render}}, I_{\mathrm{gt}})
.\end{equation}
If $S$ falls below a threshold $T_{\text{DRB}}$ (set to 0.50 in our experiments), the frame is flagged as an "exposure frame," and the DRB-Loss is activated. Otherwise, the loss remains zero, ensuring no interference under normal lighting conditions.

\textbf{DRB-Loss Calculation.} For flagged frames, we introduce a set of learnable exposure parameters $\boldsymbol{\theta} = \{g, o\}$ to apply an affine transformation to the rendered image: $I_{\mathrm{render}}^\theta = g \cdot I_{\mathrm{render}} + o$. The DRB-Loss then jointly optimizes $\boldsymbol{\theta}$ and the scene parameters:
\begin{equation}
\mathcal{L}_{\mathrm{DRB}} = (1-S) \cdot \left( \lambda_{1} \cdot \|I_{\mathrm{render}}^{\theta} - I_{\mathrm{gt}}\|_1 + \lambda_{2} \cdot \|\nabla I_{\mathrm{render}}^{\theta} - \nabla I_{\mathrm{gt}}\|_1 \right)
.\end{equation}
Crucially, the loss is weighted by $(1-S)$, making the correction adaptive: the greater the structural discrepancy (lower $S$), the stronger the applied photometric supervision. The loss combines an L1 photometric term and a gradient difference term to preserve both color and structural fidelity. The gain $g$ and offset $o$ are constrained to $[0.1, 10]$ and $[-0.2, 0.2]$ respectively, mimicking the operational range of real camera adjustments.

\subsection{Camera Tracking}
In the tracking stage, we estimate the camera pose for each new frame by leveraging the stable intrinsic map from IAN and the dynamic correction from DRB-Loss.

\textbf{Camera Pose Estimation.} We initialize the pose of the current frame $t+1$ using a constant velocity motion model: $E_{t+1} = E_t \cdot (E_t \cdot E_{t-1}^{-1})$. This initial pose is then refined by minimizing a tracking loss.

\textbf{Tracking Loss Optimization.} The camera pose is optimized by minimizing the discrepancy between rendered and observed information. The loss function combines geometric constraints (depth), appearance consistency (albedo color), and semantic alignment. The DRB-Loss is conditionally activated for frames with significant exposure issues.
\begin{equation}
\begin{aligned}
\mathcal{L}_{\mathrm{tracking}} = & \lambda_{D} \cdot \|D(p) - D_{\mathrm{gt}}\| _1 + \lambda_{C} \cdot \|C_a(p) - C_{\mathrm{gt}}\|_1 \\
& + \lambda_{S} \cdot \|S(p) - S_{\mathrm{gt}}\|_1 \\
& + \lambda_{\mathrm{DRB}} \cdot \mathbb{I}(S < T_{\mathrm{DRB}}) \cdot \mathcal{L}_{\mathrm{DRB}}.
\end{aligned}
\end{equation}
where $C_a(p)$ is the rendered intrinsic albedo, and $\mathbb{I}(\cdot)$ is the indicator function. This formulation ensures robust tracking even under challenging lighting.

\textbf{Keyframe Selection.} We employ a two-stage filtering process for keyframe selection.
First, a \textit{Geometry Filter} ensures sufficient visual overlap. We calculate the reprojection ratio $\eta$ of existing map points into the candidate frame's view:
\begin{equation}
\eta = \frac{\sum_{G_i \in G_{\mathrm{map}}} \mathbb{I}(\text{is\_in\_view}(G_i, E_{\mathrm{cand}}))}{\left|G_{\mathrm{map}}\right|}
.\end{equation}
A candidate frame is kept if $\eta$ is within a predefined range.
Second, a \textit{Semantic Filter} promotes informational diversity. We discard candidate keyframes whose rendered semantic map is identical to that of the last selected keyframe. This strategy ensures that new keyframes contribute novel semantic information, preventing redundancy and improving mapping efficiency.

\subsection{Map Construction}
In the mapping stage, we fix the camera poses of selected keyframes and jointly optimize the parameters of all Gaussian primitives ($\boldsymbol{\mu}, \boldsymbol{r}, \sigma, \boldsymbol{a}, \boldsymbol{l}, \boldsymbol{s}$).

\textbf{Joint Optimization of Map Parameters.} The mapping process is driven by a comprehensive loss function that integrates geometric, appearance, and semantic supervision from multiple keyframes. It also incorporates the IAN quantization and the conditional DRB-Loss to ensure the map's robustness and consistency.
\begin{equation}
\begin{gathered}
\mathcal{L}_{\mathrm{mapping}} = \sum_{p \in \mathcal{P}} \left( \lambda_D \cdot \|D(p) - D_{\mathrm{gt}}\| _1
+ \lambda_C \cdot \|C(p) - C_{\mathrm{gt}}\|_1 \right. \\
\left. + \lambda_S \cdot \mathcal{L}_{\mathrm{CE}}(S(p), S_{\mathrm{gt}})
+ \lambda_{\mathrm{DRB}} \cdot \mathbb{I}(S < T_{\mathrm{DRB}}) \cdot \mathcal{L}_{\mathrm{DRB}} \right).
\end{gathered}
\end{equation}
Here, $\mathcal{P}$ denotes the set of sampled pixels across keyframes. Note that for semantic supervision, we use a standard Cross-Entropy loss ($\mathcal{L}_{\mathrm{CE}}$) which is more suitable for classification tasks than SSIM. The map is dynamically densified and pruned during this process to reconstruct detailed scene geometry and appearance.

\section{EXPERIMENTS}

\begin{table*}[htbp]
  \centering
  \caption{Tracking performance on Replica\cite{replica-dataset} and ScanNet\cite{scannet}, measured by ATE RMSE (cm, lower is better $\downarrow$). Best results are in \textbf{bold}.}
  \resizebox{\linewidth}{!}{%
  \begin{tabular}{lcccccccccccccccc}
    \toprule
    Dataset & \multicolumn{8}{c}{Replica} & \multicolumn{7}{c}{ScanNet} \\
    \cmidrule(r){2-10} \cmidrule(l){11-17}
    Methods & Avg. & R0 & R1 & R2 & Of0 & Of1 & Of2 & Of3 & Of4 & Avg. & 0000 & 0059 & 0106 & 0169 & 0181 & 0207 \\
    \midrule
    Vox-Fusion\cite{Vox-fusion} & 3.09 & 1.37 & 4.70 & 1.47 & 8.48 & 2.04 & 2.58 & 1.11 & 2.94 & 26.90 & 68.84 & 24.18 & 8.41 & 27.28 & 23.30 & 9.41 \\
    NICE-SLAM\cite{Nice-slam} & 1.07 & 0.97 & 1.31 & 1.07 & 0.88 & 1.00 & 1.06 & 1.10 & 1.13 & \textbf{10.70} & 12.00 & 14.00 & \textbf{7.90} & \textbf{10.90} & 13.40 & 6.20 \\
    ESLAM\cite{eslam} & 0.63 & 0.71 & 0.70 & 0.52 & 0.57 & 0.55 & 0.58 & 0.72 & 0.63 & - & - & - & - & - & - & - \\
    Point-SLAM\cite{Point-slam} & 0.52 & 0.61 & 0.41 & 0.37 & 0.38 & 0.48 & 0.54 & 0.69 & 0.72 & 12.19 & \textbf{10.24} & \textbf{7.81} & 8.65 & 22.16 & 14.77 & 9.54 \\
    MonoGS\cite{monogs} & 0.79 & 0.47 & 0.43 & 0.31 & 0.70 & 0.57 & 0.31 & 0.31 & 3.20 & - & - & - & - & - & - & - \\
    SplaTAM\cite{splatam} & 0.36 & 0.31 & 0.40 & 0.29 & 0.47 & 0.27 & \textbf{0.29} & 0.32 & 0.55 & 11.88 & 12.83 & 10.14 & 17.72 & 12.08 & 11.10 & 7.46 \\
    SNI-SLAM\cite{Sni-slam} & 0.46 & 0.50 & 0.55 & 0.45 & 0.35 & 0.41 & 0.33 & 0.62 & 0.50 & - & - & - & - & - & - & - \\
    SemGauss-SLAM\cite{Semgauss-slam} & 0.33 & 0.26 & 0.42 & 0.27 & 0.34 & 0.17 & 0.32 & 0.36 & 0.49 & - & - & - & - & - & - & - \\
    Hier-SLAM\cite{Hier-slam} & \textbf{0.33} & \textbf{0.21} & 0.49 & \textbf{0.24} & \textbf{0.29} & \textbf{0.16} & 0.31 & 0.37 & 0.53 & 11.80 & 12.83 & 9.57 & 17.54 & 11.54 & 11.78 & 7.55 \\
    SGS-SLAM\cite{Sgs-slam} & 0.45 & 0.46 & 0.41 & 0.35 & 0.46 & 0.27 & 0.56 & 0.90 & 0.60 & 12.23 & 14.90 & 11.10 & 17.83 & 12.22 & 9.67 & 7.63 \\
    \textbf{Ours} & 0.34 & 0.25 & \textbf{0.36} & 0.33 & 0.50 & 0.21 & 0.32 & \textbf{0.30} & \textbf{0.45} & 11.30 & 14.39 & 9.07 & 17.13 & 11.91 & \textbf{8.26} & \textbf{7.06} \\
    \bottomrule
    \end{tabular}
    }
  \label{tab:ATE}
\end{table*}

\begin{table}[htbp]
  \centering
  \caption{Rendering quality on the Replica\cite{replica-dataset}. Metrics are PSNR (dB, higher is better $\uparrow$), SSIM (higher is better $\uparrow$), and LPIPS (lower is better $\downarrow$).}
  \resizebox{\linewidth}{!}{%
  \begin{tabular}{llccccccccc}
    \toprule
    Methods & Metrics & Avg. & R0 & R1 & R2 & Of0 & Of1 & Of2 & Of3 & Of4 \\
    \midrule
    \multirow{3}{*}{Vox-Fusion\cite{Vox-fusion}} & PSNR$\uparrow$ & 24.41 & 22.39 & 22.36 & 23.92 & 27.79 & 29.83 & 20.33 & 23.47 & 25.21 \\
    & SSIM$\uparrow$ & 0.80 & 0.68 & 0.75 & 0.80 & 0.86 & 0.88 & 0.79 & 0.80 & 0.85 \\
    & LPIPS$\downarrow$ & 0.24 & 0.30 & 0.27 & 0.23 & 0.24 & 0.18 & 0.24 & 0.21 & 0.20 \\
    \midrule
    \multirow{3}{*}{NICE-SLAM\cite{Nice-slam}} & PSNR$\uparrow$ & 24.42 & 22.12 & 22.47 & 24.52 & 29.07 & 30.34 & 19.66 & 22.23 & 24.96 \\
    & SSIM$\uparrow$ & 0.81 & 0.69 & 0.76 & 0.81 & 0.87 & 0.89 & 0.80 & 0.80 & 0.86 \\
    & LPIPS$\downarrow$ & 0.23 & 0.33 & 0.27 & 0.21 & 0.23 & 0.18 & 0.24 & 0.21 & 0.20 \\
    \midrule
    \multirow{3}{*}{ESLAM\cite{eslam}} & PSNR$\uparrow$ & 28.06 & 25.25 & 27.39 & 28.09 & 30.33 & 27.04 & 27.99 & 29.27 & 29.15 \\
    & SSIM$\uparrow$ & 0.92 & 0.87 & 0.89 & 0.96 & 0.93 & 0.91 & 0.94 & 0.95 & 0.95 \\
    & LPIPS$\downarrow$ & 0.26 & 0.32 & 0.30 & 0.25 & 0.21 & 0.25 & 0.24 & 0.19 & 0.21 \\
    \midrule
    \multirow{3}{*}{SplaTAM\cite{splatam}} & PSNR$\uparrow$ & 34.11 & 32.86 & 33.89 & 35.25 & 38.26 & 39.17 & 31.97 & 29.70 & 31.81 \\
    & SSIM$\uparrow$ & 0.97 & 0.98 & 0.97 & 0.98 & 0.98 & 0.98 & 0.97 & 0.95 & 0.95 \\
    & LPIPS$\downarrow$ & 0.10 & 0.07 & 0.10 & 0.08 & 0.09 & 0.09 & 0.10 & 0.12 & 0.15 \\
    \midrule
    \multirow{3}{*}{SNI-SLAM\cite{Sni-slam}} & PSNR$\uparrow$ & 29.43 & 25.91 & 28.17 & 29.15 & 31.85 & 30.34 & 29.13 & 28.75 & 30.97 \\
    & SSIM$\uparrow$ & 0.92 & 0.88 & 0.90 & 0.92 & 0.94 & 0.93 & 0.93 & 0.93 & 0.94 \\
    & LPIPS$\downarrow$ & 0.23 & 0.31 & 0.29 & 0.26 & 0.19 & 0.21 & 0.23 & 0.21 & 0.20 \\
    \midrule
    \multirow{3}{*}{SGS-SLAM\cite{Sgs-slam}} & PSNR$\uparrow$ & 34.66 & 32.50 & 34.25 & 35.10 & \textbf{38.54} & \textbf{39.20} & \textbf{32.90} & 32.90 & \textbf{32.75} \\
    & SSIM$\uparrow$ & 0.97 & 0.98 & \textbf{0.98} & 0.98 & 0.98 & 0.98 & 0.97 & \textbf{0.97} & 0.95 \\
    & LPIPS$\downarrow$ & 0.10 & 0.07 & \textbf{0.09} & 0.07 & 0.09 & 0.09 & 0.10 & 0.12 & 0.15 \\
    \midrule
    \multirow{3}{*}{\textbf{Ours}} & PSNR$\uparrow$ & \textbf{34.75} & \textbf{33.05} & \textbf{35.37} & \textbf{35.16} & 37.80 & 38.95 & 32.86 & \textbf{33.16} & 31.62 \\
    & SSIM$\uparrow$ & \textbf{0.97} & \textbf{0.98} & 0.97 & \textbf{0.98} &\textbf{ 0.98} & \textbf{0.98} & \textbf{0.97} & 0.95 & \textbf{0.95} \\
    & LPIPS$\downarrow$ & \textbf{0.10} & \textbf{0.07} & 0.10 & \textbf{0.07} & \textbf{0.08} & \textbf{0.09} & \textbf{0.10} & \textbf{0.11} & \textbf{0.15} \\
    \bottomrule
    \end{tabular}
    }
  \label{tab:Rendering Quality}
\end{table}

\begin{table}[t!]
  \centering
  \caption{Semantic segmentation performance on Replica\cite{replica-dataset}, measured by mIoU (\%, higher is better $\uparrow$).}
  \resizebox{\linewidth}{!}{%
 \begin{tabular}{lccccc}

    \toprule
    Methods & Avg. mIoU $\uparrow$ & R0 [\%] $\uparrow$ & R1 [\%] $\uparrow$ & R2 [\%] $\uparrow$ & Of0 [\%] $\uparrow$ \\
    \midrule
    NIDS-SLAM\cite{NIDS-SLAM} & 82.37 & 82.45 & 84.08 & 76.99 & 85.94 \\
    DNS-SLAM\cite{Dns-slam} & 87.77 & 88.32 & 84.90 & 81.20 & 84.66 \\
    SNI-SLAM\cite{Sni-slam} & 87.41 & 88.42 & 87.43 & 86.16 & 87.63 \\
    SGS-SLAM\cite{Sgs-slam} & \textbf{92.72} & 92.72 & \textbf{92.91} & \textbf{92.10} & \textbf{92.90} \\
    \textbf{Ours} & 92.69 & \textbf{92.78} & 92.85 & 92.02 & \textbf{93.10} \\
    \bottomrule
  \end{tabular}
}
  \label{tab:mIoU}
\end{table}

\begin{table}[t!]
  \centering
  \caption{Ablation study on the Replica\cite{replica-dataset} (Room0) using SGS-SLAM\cite{Sgs-slam} as the baseline.}
  \resizebox{\linewidth}{!}{%
 \begin{tabular}{lcccc}
    \toprule
    Settings & \makecell{Depth L1\\{[}cm{]} $\downarrow$} & \makecell{ATE RMSE\\{[}cm{]} $\downarrow$} & \makecell{PSNR\\{[}dB{]} $\uparrow$} & \makecell{mIoU\\{[}\%{]} $\uparrow$} \\
    \midrule
    Baseline(SGS-SLAM\cite{Sgs-slam}) & 0.50 & 0.46 & 32.41 & 92.69 \\
    Baseline + IAN & 0.50 & 0.36 & 32.22 & 92.40 \\
    Baseline + DRB-Loss & 0.53 & 0.30 & 32.85 & 92.56 \\
    \textbf{Ours (Full Model)} & \textbf{0.49} & \textbf{0.25} & \textbf{33.03} & \textbf{92.73} \\
    \bottomrule
  \end{tabular}
  }
  \label{tab:ablation}
\end{table}

We evaluate our method on the synthetic \textbf{Replica}\cite{replica-dataset} and real-world \textbf{ScanNet}\cite{scannet} datasets. We assess performance using established metrics for three key aspects: \textbf{rendering quality} (PSNR$\uparrow$, SSIM$\uparrow$, LPIPS$\downarrow$), \textbf{tracking accuracy} (ATE RMSE$\downarrow$), and \textbf{semantic understanding} (mIoU$\uparrow$).

\subsection{Result}
\begin{figure}[ht]
\centering
\includegraphics[scale=0.15,  trim=0cm 0cm 0cm 0cm, clip]{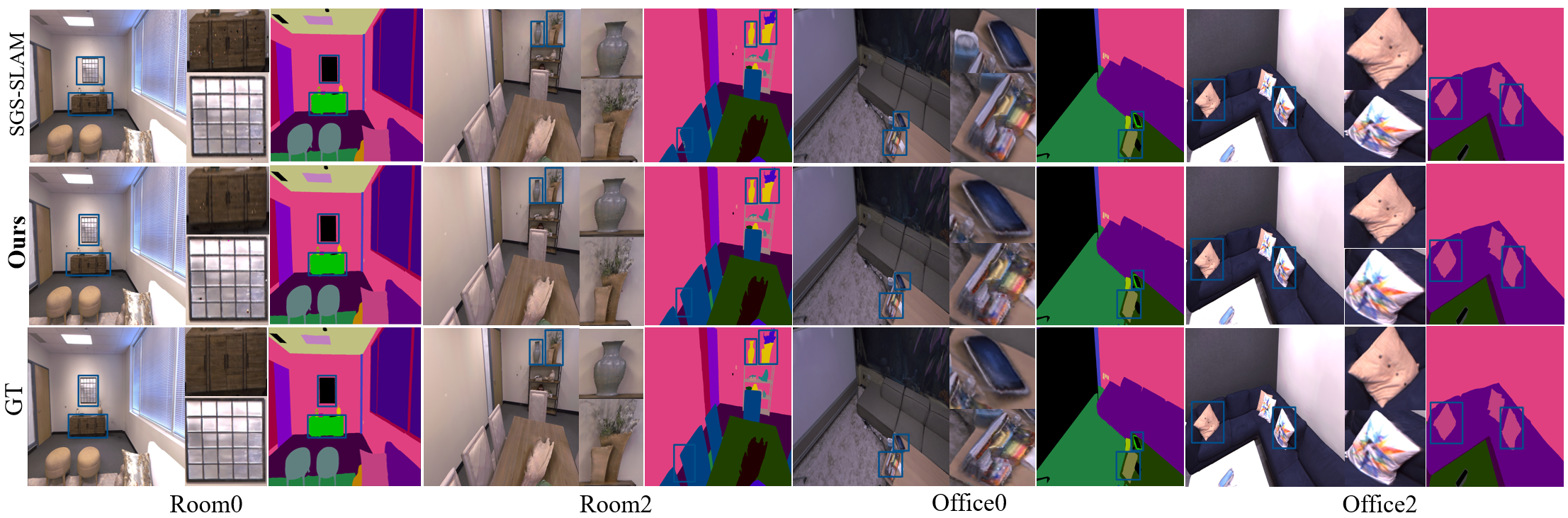}
\caption{Visualization of Our Rendering Performance on the Replica Dataset.}
\label{fig:shijue}
\end{figure}

To validate the robustness of our method against challenging illumination variations, we evaluate camera tracking accuracy on both Replica and ScanNet. As shown in Table~\ref{tab:ATE}, our method demonstrates superior tracking accuracy compared to the baseline and other state-of-the-art methods. This significant improvement is attributed to the stable intrinsic appearance representation provided by our IAN module and the dynamic compensation for exposure shifts by the DRB-Loss. (1) Specifically, our method reduces the average ATE RMSE on Replica from 0.45 cm (SGS-SLAM\cite{Sgs-slam}) to 0.34 cm and on ScanNet from 12.23 cm to 11.30 cm. Notably, our method achieves performance on par with the state-of-the-art Hier-SLAM\cite{Hier-slam} on Replica\cite{replica-dataset} and sets a new SOTA on several challenging real-world ScanNet\cite{scannet} scenes (e.g., `scene0181`, `scene0207`), proving its effectiveness and reliability in real-world conditions. (2) High-fidelity rendering is crucial for creating photorealistic digital twins. As detailed in Table~\ref{tab:Rendering Quality}, our method achieves state-of-the-art performance across all three standard metrics (PSNR, SSIM, and LPIPS) on average. This demonstrates that our framework not only reconstructs accurate geometry but also learns a consistent and true-to-life scene appearance, effectively mitigating artifacts such as color inconsistencies and detail loss that are commonly caused by lighting changes. The superior rendering quality is also visually confirmed in Fig.~\ref{fig:shijue}. (3) Our method also excels in the downstream task of semantic segmentation. As shown in Table~\ref{tab:mIoU}, our system achieves a top-tier mIoU of 92.69\%, which is on par with the state-of-the-art SGS-SLAM (92.72\%) while surpassing it in several scenes (e.g., R0, Of0). This result underscores a key tenet of our work: a robust geometric and appearance foundation, invariant to lighting, provides a clean and stable input for semantic feature extraction. This prevents misclassifications that can arise from photometric inconsistencies, thereby ensuring high performance in scene understanding tasks.

\subsection{Ablation Studies}
We conduct a thorough ablation study on the Replica\cite{replica-dataset} to dissect the contribution of our two core components: the proactive Intrinsic Appearance Normalization (IAN) module and the reactive Dynamic Radiance Balancing Loss (DRB-Loss). The results, presented in Table~\ref{tab:ablation}, clearly validate our design choices. (1) Effect of IAN: Adding only the IAN module to the baseline dramatically reduces the ATE RMSE from 0.46 cm to 0.36 cm. This provides strong evidence that regularizing the appearance via our quantization scheme yields a stable intrinsic representation that is crucial for robust tracking under varying illumination.(2)Effect of DRB-Loss: Individually, the DRB-Loss module also significantly improves tracking accuracy (ATE to 0.30 cm) while simultaneously boosting the rendering quality (PSNR from 32.41 to 32.85 dB). This highlights its dual benefit in handling extreme exposure frames for both tracking and mapping. (3) Full Model: Finally, our full model, integrating both modules, achieves the best performance across the board, reaching the lowest tracking error (0.25 cm) and the highest rendering quality and semantic accuracy. This demonstrates a clear synergistic effect between our proactive normalization (IAN) and reactive compensation (DRB-Loss) strategies, which work together to create a highly robust SLAM system. The qualitative effect of the IAN module is visualized in Fig.~\ref{fig:INA}.

\vspace{-18pt}
\section{conclusion}
\vspace{-9pt}
In this work, we introduced an illumination-invariant semantic 3DGS-SLAM framework that effectively "tames the light" in real-world scenes. Our core strategy combines a proactive \textbf{Intrinsic Appearance Normalization (IAN)} module, which learns a canonical albedo, with a reactive \textbf{Dynamic Radiance Balancing (DRB) Loss} to compensate for exposure shifts. Extensive experiments and ablation studies validate our approach, demonstrating state-of-the-art performance in tracking, rendering, and semantic segmentation. By successfully disentangling intrinsic scene properties from transient lighting, our work marks a significant step towards deploying robust 3DGS-SLAM systems for real-world applications like robotics and augmented reality.

\bibliographystyle{IEEEtran}
\bibliography{refs}

\end{document}